\documentclass[letterpaper, 10 pt, conference]{ieeeconf}  

\IEEEoverridecommandlockouts                              

\usepackage{graphics} 
\usepackage{epsfig} 
\usepackage{mathptmx} 
\usepackage{times} 
\usepackage{amsmath} 
\usepackage{amssymb}  
\usepackage{algorithm,algorithmic}
\usepackage{multirow}
\usepackage{epstopdf}
\usepackage{url}
\usepackage[hidelinks]{hyperref}
\usepackage{float}
\newcommand{\MG}{\mathcal{G}}
\newcommand{\MD}{\mathcal{D}}
\newcommand{\ME}{\mathcal{E}}
\newcommand{\MS}{\mathcal{S}}
\newcommand{\MU}{\mathcal{U}}

\newcommand{\argmin}{\mathop{\rm argmin}}
\usepackage{color}

\title{\LARGE \bf
An Efficient Index for Visual Search in Appearance-based SLAM}

\author{Kiana Hajebi and Hong Zhang
\thanks{K. Hajebi and H. Zhang are with Faculty of Computing Science,
        University of Alberta, Canada,
        {\tt\small [hajebi, hzhang] at ualberta.ca}}%
}

\begin{document}

\maketitle
\thispagestyle{empty}
\pagestyle{empty}

\begin{abstract}
Vector-quantization can be a computationally expensive step in visual bag-of-words (BoW) search when the vocabulary is large. A BoW-based appearance SLAM needs to tackle this problem for an efficient real-time operation. We propose an effective method to speed up the vector quantization process in BoW-based visual SLAM. We employ a graph-based nearest neighbor search (GNNS) algorithm to this aim, and experimentally show that it can outperform the state-of-the-art. The graph-based search structure used in GNNS can efficiently be integrated into the BoW model and the SLAM framework. 
The graph-based index, which is a $k$-NN graph, is built over the vocabulary words and can be extracted from the BoW's vocabulary construction procedure, 
by adding one iteration to the k-means clustering, which adds small extra cost.
Moreover, exploiting the fact that images acquired for appearance-based SLAM are sequential, GNNS search can be initiated judiciously which helps increase the speedup of the quantization process considerably.

\end{abstract}


\section{INTRODUCTION}

Bag-of-Words (BoW) method was originally proposed for document retrieval. In recent years, the method has been successfully applied to image retrieval tasks in computer vision community \cite{nister}, \cite{Sivic2}. The method is attractive because of its efficient image representation and retrieval. BoW represents an image as a sparse vector of visual words, and thus images can be searched efficiently using an inverted index file system. Moreover, because the complexity of BoW does not grow with the size of the dataset, as much as that of other search techniques (e.g., direct feature matching) do, it can be employed for large-scale image search applications.

One major application area that benefits from BoW is the appearance-based mobile robot localization and SLAM\footnote{Simultaneously Localization And Mapping}. SLAM employs BoW to solve the \emph{loop closure detection} (LCD) problem which is a classic and difficult problem in SLAM. LCD is addressed as a place recognition problem: robot should be able to recognize places it has visited before to localise itself or refine the map of the environment. This task is performed by matching the current view of the robot to the existing map that contains the images of the previously visited locations.

In large-scale environments, SLAM maps contain a large number of images to match in order to solve the loop closure detection problem. The image search in such large maps is challenging and still an open problem. Although BoW proposes an efficient search technique, its vector quantization (VQ) step can be computationally expensive. Vector quantization maps the image feature descriptors to the words in a visual vocabulary. Typically hundreds to thousands of features are extracted from an image and need to be matched against tens or hundreds of thousands of visual words. Approximate nearest neighbor search algorithms, e.g. hierarchical k-means tree \cite{nister} and randomized KD-trees \cite{RKDtree_Silpa-AnanH08}, have been used to speed up the quantization process, however at the cost of search accuracy. 
In this paper, we adapt the graph-based nearest neighbor search algorithm (GNNS) proposed in \cite{Hajebi} to increase the efficiency of the vector quantization. GNNS utilizes a $k$-NN graph as the search index, which is constructed in an offline pre-processing phase. However, the graph construction can be expensive when the dataset (i.e., the vocabulary in our application) is large. Fortunately by integrating GNNS into the BoW model, the $k$-NN graph can be extracted from the k-means procedure,
employed for visual vocabulary construction in BoW-based algorithms, and it just adds small additional computational cost compared to the cost of k-means clustering. Most importantly, we show that GNNS can exploit the sequential dependency in SLAM data to speed up the vector quantization process considerably unlike other search indices in which it is difficult to exploit such sequential dependency in data. This motivates the use of GNNS rather than other search algorithms in solving loop closure detection problem.

To support our claim, we experimentally show that adapting GNNS to solve the vector quantization problem, 
outperforms the state-of-the-art search methods by decreasing the number of distance computations performed and hence increasing the search speedup,

\section{BACKGROUND}
\label{sec:bakgnd}

\subsection{Bag-of-Words for Image Retrieval}
\label{sec:bow}

Bag-of-words is a popular model that has been used in image classification, objection recognition, and appearance-based navigation. Because of its simplicity and search efficiency it has also been used as a successful method in large-scale image and document retrieval \cite{Sivic:2003,nister,Sivic2}. 

Bag-of-words model represents an image by a sparse vector of visual words. Image features, e.g., SIFTs \cite{Lowe:2004}, are sampled and clustered (e.g., using \emph{k-means}) in order to quantize the space into a discrete set of visual words. The centroids of clusters are then considered as visual words which form the visual vocabulary. During image retrieval or robot navigation, when a new image arrives, its local features are extracted and vector-quantized into the visual words. Each word might be weighed by some score which is either the word frequency in the image (i.e., \emph{tf}) or the ``term frequency-inverse document frequency'' or \emph{tf-idf} \cite{Sivic:2003}. A histogram of weighted visual words, which is typically sparse, is then built and used to represent the image.

An inverted index file, used in the BoW framework, is an efficient image search tool in which the visual words are mapped to the database images. Each visual word serves as a table index and points to the indices of the database images in which the word occurs. Since not every image contains every word and also each word does not occur in every image, the retrieval through inverted-index file is independent of the map size and therefore fast.


\subsection{Bag-of-Words for Appearance-based SLAM}
\label{sec:relwork}

Bag-of-words model has been extensively used as the basis of the image search in appearance-based localization or SLAM algorithms \cite{fabmap,fabmap2,AcharJK11,Angeli08,bi-BoW}.
Cummins and Newman \cite{fabmap,fabmap2} propose a probabilistic framework over the bag-of-words representation of locations, for the appearance-based place recognition. Along with the visual vocabulary they also learn the Chow Liu tree to capture the co-occurrences of visual words. Similarly, Angeli et al. \cite{Angeli08} develop a probabilistic approach for place recognition in SLAM. They build two visual vocabularies incrementally and use two BoW representations as an input of a Bayesian filtering framework to estimate the likelihood of loop closures.

Assuming each image has hundreds of features, mapping the features to the visual words, using a linear search method, is computationally prohibitive and not practical for real-time localization. Researchers have tackled this problem with different approaches that speed up the search but at the expense of accuracy. A number of papers have employed compact feature descriptors that speed up the search. G\'{a}lvez-L\'{o}pez and Tard\'{o}s in \cite{bi-BoW} propose to use FAST \cite{FAST} and BREIF \cite{BRIEF} binary features and introduce a BoW model that descritizes a binary space. Similarly \cite{fabmap,fabmap2} use SURF \cite{Surf} to have a more compact feature descriptor. Another approach is to use approximate nearest neighbor search algorithms, like hierarchical k-means \cite{nister}, KD-trees \cite{fabmap} or locality sensitive hashing (LSH) \cite{LSH_shahbazi}, to speed up the quantization process.
Li and Kosecka \cite{Kosecka} and Zhang \cite{BoRF} propose reducing the number of features in each image, thereby reducing or removing the vector-quantization process.



\section{Search Indices for Vector Quantization}
Vector-quantization as a nearest-neighbor search classification maps the high-dimensional feature descriptors into visual words. When the vocabulary is large, in order to provide sufficient discriminating power for image matching, the vector-quantization process can be a computationally expensive task. An efficient approximate nearest-neighbor search method is therefore required to speed up the search while minimizing the impact of the search method on accuracy. 

In the following two subsections, we briefly explain two of the most popular approximate nearest neighbor search methods that are widely-adapted to BoW search: hierarchical k-means and KD-trees. Our method will be compared against these two methods in the evaluation section.

\subsection{KD-trees}

The classical KD-tree algorithm \cite{Bentley:1975,kdtree2,kdtree1} partitions the space by hyper-planes that are perpendicular to the coordinate axes. At the root of the tree a hyperplane orthogonal to one of the dimensions splits the data into two halves according to some splitting value, which is usually the median or the mean of the data to be partitioned. Each half is recursively partitioned into two halves with a hyperplane through a different dimension. Partitioning stops after $\log n$ levels, where n is the total number of data points, so that at the bottom of the tree each leaf node corresponds to one of the data points. The splitting values at each level are stored in the nodes. The query point is then compared to the splitting value at each node while traversing the tree from root to leaf to find the nearest neighbor. Since the leaf point is not necessarily the nearest neighbor, to find approximate nearest neighbors, a backtrack step from the leaf node is performed and the points that are closer to the query point in the tree are examined. Instead of simple backtracking, \cite{BBFKDtree} proposes to use Best-Bin-First (BBF) heuristic to perform the search faster. In BBF one maintains a sorted queue of nodes that have been visited and expands the bins in the order of their distance to the query point (\emph{priority search}).

\subsection{Hierarchical K-means tree}

Hierarchical k-means trees (HKM), proposed by \cite{HKM_Fukunage:1975}, is another type of partitioning trees based on k-means clustering. The tree is built by running k-means clustering recursively. The data points are first partitioned into $k$ distinct clusters to form the nodes of the first layer of the tree. Each cluster is recursively partitioned into $k$ (called branching factor) clusters and this process continues until there is no more than $k$ data points in each node. A depth-first search is a common tree traversal approach for searching the tree. \cite{FLANN} proposes to use priority queues to search the tree more efficiently. Similar to the Best-Bin-First approach \cite{BBFKDtree}, when traversing the tree, the unvisited branches of the nodes along the path, are added to the priority queue. When backtracking, the branches, in the order of their distance (i.e. the distance of their cluster centroid) to the query, are extracted and expanded.

\section{PROPOSED METHOD}

Our fast vector quantization method employs a graph-based $k$-nearest neighbor search algorithm, called GNNS, that outperforms the popular ANN-search methods widely-used in BoW models and SLAM systems. In the following sub-sections we describe GNNS's properties and how nicely GNNS can be adapted for the vector quantization in BoW and SLAM system.

\subsection{The Graph Nearest Neighbor Search (GNNS)}

The graph nearest neighbor search (GNNS) algorithm used in this work, has been originally proposed in \cite{Sebastian:Kimia} and independently re-discovered in \cite{Hajebi}. GNNS builds a $k$-NN graph and, when queried with a new point, it performs hill-climbing starting from a randomly sampled node of the graph. In our application the graph is constructed in an offline phase, which is explained in the following sub-sections\footnote{The following sub-sections are re-stated from our paper \cite{Hajebi} for the clarity of the presentation}. 

\subsection{$k$-NN Graph Construction}
\label{sec:KNNG}

A $k$-NN graph is a directed graph $\MG=(\MD,\ME)$, where $\MD$ is the set of nodes (i.e. datapoints) and $\ME$ is the set of links. Node $X_i$ is connected to node $X_j$ if $X_j$ is one of the $k$-NNs of $X_i$. The computational complexity of the naive construction of this graph is $O(dn^2)$, where $n$ is the size of the dataset and $d$ is the dimensionality.

The choice of $k$ is crucial to have a good performance. A small $k$ makes the graph too sparse or disconnected so that the hill-climbing method frequently gets stuck in local minima. Choosing a big $k$ gives more flexibility during the runtime, but consumes more memory and makes the offline graph construction more expensive.

\subsection{Approximate $K$-Nearest Neighbor Search Algorithm}
\label{sec:GNNS}

\begin{table}
\begin{center}
\framebox{\parbox{8cm}{  
\begin{algorithmic}
\STATE \textbf{Input}: a $k$-NN graph $\MG=(\MD,\ME)$, a query point $Q$, the number of nearest neighbors to be returned $K$, the number of random restarts $R$, the number of greedy steps $T$, and the number of expansions $E$.
\STATE $\rho$ is a distance function. $N(Y, E, \MG)$ returns the first $E$ neighbors of node $Y$ in $\MG$.
\STATE $\MS=\{\}$.
\STATE $\MU=\{\}$.
\STATE $Z=X_1$.
\FOR{$r=1,\dots, R$}
	\STATE $Y_0$: a point drawn randomly from a uniform distribution over $\MD$.
	\FOR{$t=1,\dots,T$}
		\STATE $Y_{t}=\argmin_{Y\in N(Y_{t-1},E,\MG)} \rho(Y,Q)$.
		\STATE $\MS=\MS\bigcup N(Y_{t-1},E,\MG)$.
		\STATE $\MU=\MU\bigcup\{\rho(Y,Q):\ Y\in N(Y_{t-1},E,\MG)\}$.
	\ENDFOR
\ENDFOR
\STATE Sort $\MU$, pick the first $K$ elements, and return the corresponding elements in $\MS$.
\end{algorithmic}
}}
\end{center}
\caption{The Graph Nearest Neighbor Search (GNNS) algorithm for $K$-NN Search Problems.}
\label{alg:GNNS}
\end{table}

\begin{figure}[tp]
\centering
\includegraphics[width=2in]{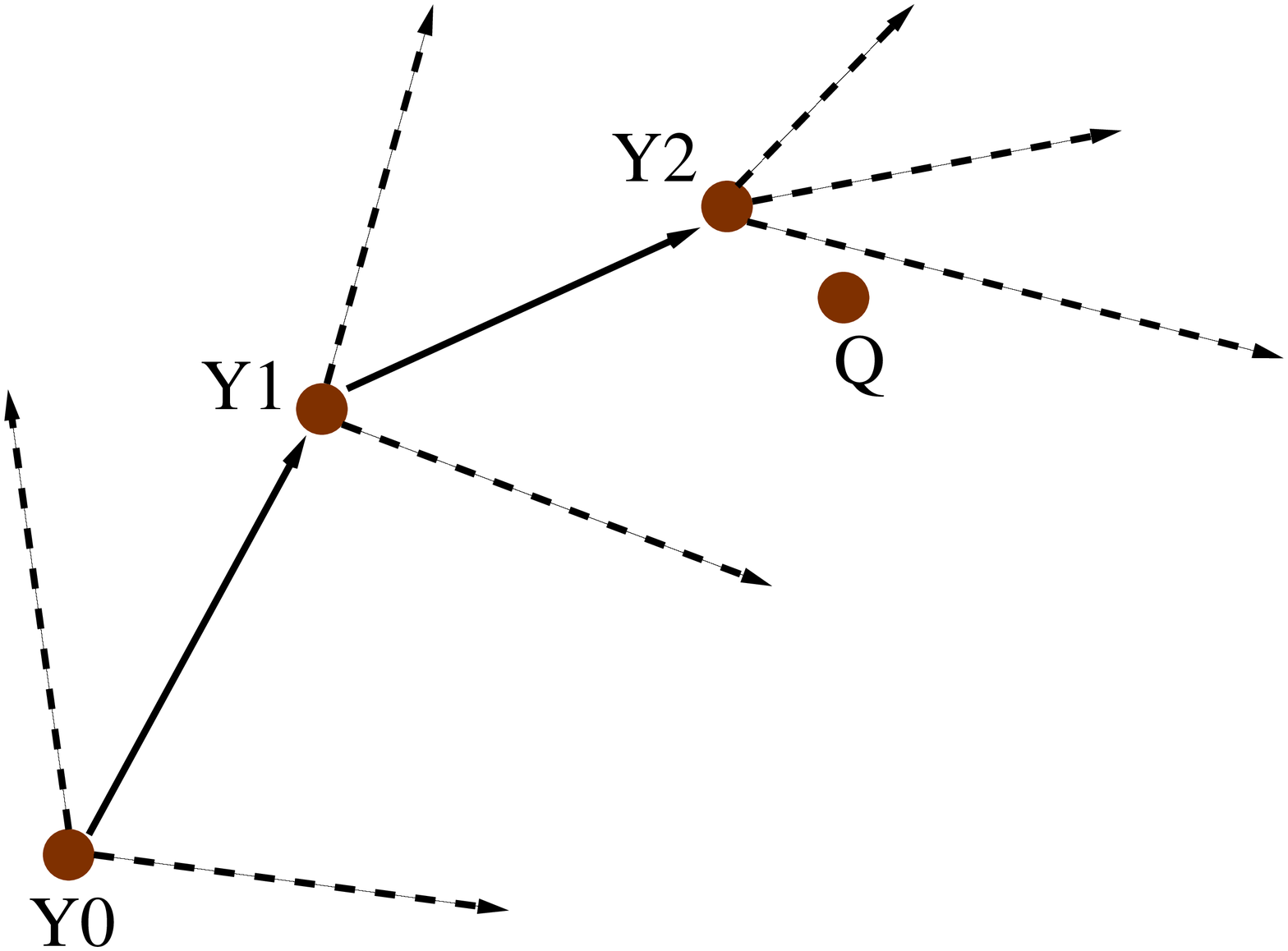}
\caption{The GNNS Algorithm on a simple nearest neighbor graph.}
\label{fig:NNG}
\end{figure}

The GNNS algorithm, which is basically a best-first search method to solve the $K$ nearest neighbor search problem, is shown in Table~\ref{alg:GNNS}. Throughout this section, we use capital $K$ to indicate the number of nearest neighbors to be returned
, and small $k$ to indicate the number of neighbors to each node in the $k$-nearest neighbor graph.
Starting from a randomly chosen node from the $k$-NN graph, the algorithm replaces the current node $Y_{t-1}$ by the neighbor that is closest to the query:
$$Y_t=\argmin_{Y\in N(Y_{t-1},E,\MG)} \rho(Y,Q),$$
where $N(Y,E,\MG)$ returns the first $E\leq k$ neighbors of $Y$ in $\MG$, and $\rho$ is a distance measure (we use Euclidean distance in our experiments).
The algorithm terminates after a fixed number of greedy moves $T$. 
We can alternatively terminate the search when the algorithm reaches a node that is closer to the query than its best neighbor. At termination, the current best $K$ nodes are returned as the $K$ nearest neighbors to the query. Figure~\ref{fig:NNG} illustrates the algorithm on a simple nearest neighbor graph with query $Q$, $K=1$ and $E=3$.

Parameters $R$, $T$, and $E$ specify the computational budget of the algorithm. By increasing each of them, the algorithm spends more time in search and returns a more accurate result. The difference between $E$ and $k$ and $K$ should be noted.
$E$ and $K$ are two input parameters to the search  algorithm (online), while $k$ is a parameter of the $k$-NN graph construction algorithm (offline).
Given a query point $Q$, the search algorithm has to find the $K$ nearest neighbors of $Q$. The algorithm, in each  greedy  step,  examines  only  $E$  out  of  $k$  neighbors  (of  the  current  node)  to  choose  the  next  node. Hence, it effectively works on an $E$-NN graph.

In our vector quantization application in this paper, we set $K=1$ as we do not need to assign more than one visual word to each image feature. 

\subsection{Graph Vocabulary Construction}
The vocabulary of a BoW model is usually constructed using k-means clustering. Feature descriptors from a training dataset are first clustered into visual words, and then a search index is built over the visual words in an offline phase. 

In the previous sub-section, we mentioned to use a $k$-NN graph as the search index for the vocabulary so that each visual word corresponds to a node in the graph. However, the graph construction is a computationally expensive process especially when the dataset is large. Even though this search index is constructed only once offline, it is still desirable to minimize its computational cost.

As an alternative, we show that the graph construction can be efficiently integrated into the k-means clustering process. 
In k-means processing, in every iteration, the distance of data points to each cluster centroid is computed to update their membership in the new clusters, and this continues until convergence. In the end, the cluster centroids are chosen as visual words. Given $n$ data points and $C$ clusters centroids\footnote{The number of clusters generated by k-means is generally denoted by ``k'', however to avoid confusion with other $k$ and $K$ notations we use for $k$-NN graph and $K$-NN search, we use $C$ to denote the number of cluster centroids.}, the complexity of each iteration is then $O(nC)$. The $k$-NN graph construction, can be embedded into k-means processing as follows: 
in the last iteration, in addition to other data points, the distance of each centroid from the other centroids is computed and the nearest neighbors are used to build the $k$-NN graph. The additional computational cost is negligible, as $C\ll n$. The complexity of the last iteration will then change to $O((n+C)C)$ which is slightly higher than $O(nC)$. This shows that the complexity of graph construction is comparable to the complexity of one iteration of k-means clustering and in applications where k-means clustering is performed (like in BoW for vocabulary construction), the construction time of graph-based index is absorbed by the k-means algorithm.

However, k-means clustering with the time complexity of $O(nC)$ is an expensive process and only practical for applications that require small vocabularies ($C < 10^5$). The hierarchical k-means (HKM) \cite{nister} with complexity $O(n\log C)$ has been proposed to reduce the computational cost of k-means. Philbin et. al [25] propose a modification of k-means where the exact nearest neighbor search is replaced by an approximate NN-search algorithm, e.g. KD-trees. They show that this modification achieves the complexity of HKM. They also demonstrate that approximate k-means (AKM) outperforms HKM, when applied to the vector quantization problem. This justifies the use of k-means or approximate k-means clustering in generating the visual vocabulary (before graph construction) in our proposed method.

\subsection{Exploiting Sequential Dependencies in Data}
\label{sec:seq}
The sequential property of data can be utilized to the advantage of image retrieval. 
Unlike many image retrieval and classification applications that search in a pool of unordered images, in appearance-based SLAM we can take advantage of the temporal coherency of images to make the image search for loop closure detection efficient. We proposed to use GNNS to do the vector-quantization. In standard GNNS, search is initiated from a random node. By taking the sequential property of images into account, we can make the random initiation of GNNS smarter: sequential images usually have considerable overlap with their neighboring images and hence share a certain amount of features and visual words. 
This property can reduce the amount of computations required for the vector-quantization step, as we can quantize a feature once in the image where it is first observed, and use its visual word in subsequent images as long as the feature is observed. This requires us to match each new frame to the previous frame(s) to find the repeatable features. This step does not incur a significant cost as feature matching in two images can be performed efficiently and can in fact incur no additional costs if the matching between consequent frames is already done in the process of key-frame detection: a new frame is matched against the previous key-frame in order to decide whether it is sufficiently different from the previous key-frame in terms of appearance to be considered a new location in the map. This process is done through direct feature matching between images \cite{keyframe}.

Our approach works as follows: once a new image is captured, the features are extracted. Each feature is vector-quantized through GNNS. Let $f$ be a feature in the current image that has a match $f^\prime$ in the previous key-frame. Let the visual word assigned to $f^\prime$ be $w^\prime$. Intuitively, there is a good chance that the visual word $w^\prime$ is also the word or one of the neighbors of the word corresponding to feature $f$. Therefore we start the GNNS search from $w^\prime$, rather than a random node. This can significantly reduce the number of iterations and distance computations in the GNNS search. This is an advantage of the graph-based index over other search indices when images to be processed are temporally dependent as in visual SLAM, as it is not trivial to employ such prior knowledge in other search indices.

In the GNNS algorithm described in Section \ref{sec:GNNS}, $R$, which indicates the number of random restarts, can take any number based on our computational budget. However, our proposed sequential method helps us choose only one good initial node to start the search with. Having $R>1$ might reduce the efficiency of our sequential algorithm. So we set $R$ to 1 in our experiments. As we show in the experiments section, this is still an effective choice in practice. In addition, we choose the version of GNNS in which the search terminates at the local minima, instead of having $T$ greedy moves. Note that improving the speedup, is not possible if we always make a fixed number of greedy moves.

\section{EXPERIMENTS}
In this section, we compare the performance of our method with hierarchical k-means (HKM) and KD-trees, when applied to the problem of vector quantization in the context of visual SLAM. We will describe the datasets we used for performance evaluation of all methods, followed by discussion of our experimental results.

\subsection{Datasets}
We performed our experiments on two datasets: an outdoor and an indoor dataset. The outdoor dataset is the City Center dataset\footnote{\url{http://www.robots.ox.ac.uk/~mobile/IJRR_2008_Dataset/}} (left-side sequence) \cite{fabmap} and contains 1237 images. The indoor dataset is a \emph{lab} dataset that has been taken inside a research laboratory using an ActivMedia Pioneer P3-AT mobile robot equipped with a Dragonfly IEEE-1394 camera, and contains X images. 
Two different vocabularies with different sizes, 5000-word and 204,000-word, have been used for our study, that have been constructed using k-means clustering. We clustered $128$-dimensional SIFT \cite{Lowe:2004} feature descriptors extracted from different datasets than the above-mentioned. The 204,000-word one is used to evaluate the performance of our method on large-scale data.

\subsection{Vector Quantization Performance Evaluation}

We compare the performance of four methods on vector quantization. Randomized KD-trees, hierarchical k-means tree (HKM), GNNS and our proposed method, Sequential GNNS (SGNNS), which is the GNNS that considers the sequential property of data. For the KD-trees and HKM we used FLANN library\footnote{[\url{http://www.cs.ubc.ca/~mariusm/index.php/FLANN/FLANN}]} implementations.

Tables \ref{table:exp1}-\ref{table:exp4} compare the results of four search algorithms on vector quantization. The first two columns show the results when all image features are quantized, and the last two columns show the results when only matched features between images are quantized. 
Matched features between each image and the previous keyframe are detected using feature matching with distance-ratio test and epipolar geometric verification. In SGNNS's implementation, for the features that have a match in the previous frame, GNNS search starts from the visual word assigned to their match in the previous frame. For the rest of image features, GNNS starts from a random node.

Our performance evaluation metric is the speedup over linear search while fixing the search accuracy in terms of the true nearest neighbors found. We select the parameters in the algorithms such that we can obtain a fixed accuracy and then calculate the speedup of the algorithms over the linear search at the same accuracy. The speedup over linear search is computed as the ratio of the number of distance computations one algorithm performs over the number of distance computations linear search performs.

The only parameter in GNNS and SGNNS that we set here is $k$, the size of the $k$-NN graph index. We explained in Section \ref{sec:seq} and \ref{sec:GNNS} how the other parameters are chosen. $E$, the number of expansions is also equal to $k$ in our experiments.

In the case of KD-trees, the FLANN parameters that we set include \emph{trees}, the number of randomized KD-trees, and \emph{checks}, the number of leaf nodes to check in one search. In the case of HKM, the parameter set includes \emph{iterations}, the maximum number of iterations to perform in one k-means clustering, \emph{branching}, which is the branching factor of the tree, and \emph{checks}, the number of leaf nodes to check.

Tables \ref{table:exp1}-\ref{table:exp3} shows the experiments on City Center dataset using the 5000-word vocabulary. 
The average number of SIFT features extracted from each image is 316 and the average number of matched features is 42, which is roughly 13\% of all features. The speedups for accuracies of $\sim$87\% and $\sim$99\% have been shown in the first two tables.

To obtain the results in Table \ref{table:exp1}, we used a 50-NN graph for GNNS and SGNNS. For KD-trees, we set \emph{trees} and \emph{checks} to 1 and 200, respectively. For HKM, we set \emph{iterations}, \emph{branching} and \emph{checks} to 3, 8 and 40, respectively.

In Table \ref{table:exp2}, we used a 200-NN graph for GNNS and SGNNS. For KD-trees, we set \emph{trees} and \emph{checks} to 4 and 600, respectively. For HKM, we set \emph{iterations}, \emph{branching} and \emph{checks} to 7, 8 and 160, respectively.

In Table \ref{table:exp3}, we show the vector quantization results when the 204,000-word vocabulary is used. We used a 300-NN graph for GNNS and SGNNS. For KD-trees, we set \emph{trees} and \emph{checks} to 4 and 2200, respectively. For HKM, we set \emph{iterations}, \emph{branching} and \emph{checks} to 7, 8 and 500, respectively.

The last part of experiments has been done on an indoor dataset (Table \ref{table:exp4}) where the overlap between images is larger (around 82\% of features are matched). We used the 5000-word vocabulary for this experiment. For GNNS and SGNNS we used a 250-NN graph and for KD-trees, we set \emph{trees} and \emph{checks} to 6 and 400, respectively. For HKM, we set \emph{iterations}, \emph{branching} and \emph{checks} to 3, 8 and 160, respectively. 

As can be seen in all experiments, both GNNS and HKM outperform KD-trees, and SGNNS, our proposed method, outperforms HKM by as much as 50\% for the case of vector-quantizing all features, or as much as 300\% if only matched features are used in creating the BoW representation of an image. This superior performance is due to both the efficiency of graph-based search (GNNS) - as indicated by the third row of each table, over the first two rows of each table - and by the exploitation of the sequential property of images whose features are to be vector-quantized, as indicated by the last row (SGNNS) of each table over the third row (GNNS).


We also observed that the features that are common between two images do not share common visual words, necessarily. On average, 64\% of corresponding features share the same visual words, in the experiments presented in Tables \ref{table:exp1}-\ref{table:exp2}. This amount was reduced to 36\%, when we used the 204k vocabulary (Table \ref{table:exp3}).

\begin{table}[!h]
\caption{Comparison of different search algorithms on Vector Quantization - City Center dataset, Accuracy fixed at $\sim$87\%}
\label{table:exp1}
\begin{center}
\tabcolsep 4.5pt
{\renewcommand{\arraystretch}{1.2}
\begin{tabular}{|c|c|c||c|c|}
  \cline{2-5}
  \multicolumn{1}{c}{} & \multicolumn{2}{|c||}{VQ of all features} & \multicolumn{2}{c|}{VQ of matched features} \\  \cline{2-5}
  \multicolumn{1}{c|}{} & Accuracy & Speedup & Accuracy & Speedup \\  \hline

  KD & 0.8839 & 8.4527 & 0.9209 & 8.3847  \\ \hline
  HKM & 0.8892 & 27.0361 & 0.9130 & 26.1947  \\ \hline
  GNNS & 0.8665 & 23.3501 & 0.8904 & 23.8412  \\ \hline
  SGNNS & 0.8783 & 32.1414 & 0.9624 & 81.5666  \\ \hline
\end{tabular}}
\end{center}

\begin{tabular}{cc}

 \vspace{.1 cm}
\end{tabular}

\caption{Comparison of different search algorithms on Vector Quantization - City Center dataset, Accuracy fixed at $\sim$99\%} 
\label{table:exp2}
\begin{center}
\tabcolsep 4.5pt
{\renewcommand{\arraystretch}{1.2}
\begin{tabular}{|c|c|c||c|c|}
  \cline{2-5}
  \multicolumn{1}{c}{} & \multicolumn{2}{|c||}{VQ of all features} & \multicolumn{2}{c|}{VQ of matched features} \\  \cline{2-5}
  \multicolumn{1}{c|}{} & Accuracy & Speedup  & Accuracy & Speedup \\  \hline
  KD & 0.9899 & 2.0767 & 0.9951 & 2.0678  \\ \hline
  HKM & 0.9925 & 6.9204 & 0.9956 & 6.3487  \\ \hline
  GNNS & 0.9886 & 7.4055 & 0.9917 & 7.5927  \\ \hline
  SGNNS & 0.9893 & 9.3872 & 0.9952 & 20.5244  \\ \hline
\end{tabular}}
\end{center}

\begin{tabular}{cc}

 \vspace{.1 cm}
\end{tabular}

\caption{Comparison of different search algorithms on Vector Quantization - City Center dataset, Accuracy fixed at $\sim$92\%} 
\label{table:exp3}
\begin{center}
\tabcolsep 4.5pt
{\renewcommand{\arraystretch}{1.2}
\begin{tabular}{|c|c|c||c|c|}
  \cline{2-5}
  \multicolumn{1}{c}{} & \multicolumn{2}{|c||}{VQ of all features} & \multicolumn{2}{c|}{VQ of matched features} \\  \cline{2-5}
  \multicolumn{1}{c|}{} & Accuracy & Speedup  & Accuracy & Speedup \\  \hline
  KD & 0.9319 & 22.4525 & 0.9376 & 22.3318  \\ \hline
  HKM & 0.9024 & 86.7133 & 0.8984 & 88.0266  \\ \hline
  GNNS & 0.9221 & 90.7018 & 0.9329 & 88.9622  \\ \hline
  SGNNS & 0.9236 & 121.7556 & 0.9401 & 285.3466  \\ \hline
\end{tabular}}
\end{center}

\begin{tabular}{cc}

\vspace{.1 cm}
\end{tabular}

\caption{Comparison of different search algorithms on Vector Quantization - Lab dataset, Accuracy fixed at $\sim$98\% } 
\label{table:exp4}
\begin{center}
\tabcolsep 4.5pt
{\renewcommand{\arraystretch}{1.2}
\begin{tabular}{|c|c|c||c|c|}
  \cline{2-5}
  \multicolumn{1}{c}{} & \multicolumn{2}{|c||}{VQ of all features} & \multicolumn{2}{c|}{VQ of matched features} \\  \cline{2-5}
  \multicolumn{1}{c|}{} & Accuracy & Speedup  & Accuracy & Speedup \\  \hline
  KD & 0.9833 & 2.9367 & 0.9829 & 2.9369  \\ \hline
  HKM & 0.9921 & 7.2662 & 0.9919 & 7.2633  \\ \hline
  GNNS & 0.9886 & 6.0819 & 0.9894 & 6.0555  \\ \hline
  SGNNS & 0.9935 & 15.8861 & 0.9955 & 18.3208  \\ \hline
\end{tabular}}
\end{center}

\end{table}

\section{CONCLUSION}

We proposed to use the graph nearest search (GNNS) algorithm to speed up the vector quantization task in BoW. We described the GNNS's properties and how it can integrated into the BoW and SLAM framework, taking advantage of the sequential property of SLAM data. We experimentally showed significant improvements over the state-of-the-art algorithms. We also observed that there is a bigger improvement if we only consider the matched features. 

\bibliography{icra}

\end{document}